\documentclass[10pt,conference]{IEEEtran}

\usepackage{subimages}
\usepackage[utf8]{inputenc}
\usepackage[T1]{fontenc}
\usepackage{verbatim}
\usepackage{cite, color}
\usepackage{multirow}
\usepackage{amssymb}
\usepackage{float}
\usepackage{graphicx}
\usepackage{amsthm,amsmath, mathtools}
\usepackage{cleveref}
\usepackage[ruled,vlined,linesnumbered]{algorithm2e}
\usepackage{url,bm}
\usepackage{graphicx}
\usepackage{placeins}
\usepackage{booktabs}

\Crefname{subsection}{Subsection}{Subsections}

\title{Video Segmentation Learning Using Cascade Residual Convolutional Neural Network}

\newif\iffinal
\finaltrue

\iffinal
  \author{%
    \IEEEauthorblockN{Daniel~F.~S.~Santos$^{\dag}$\thanks{$^{\dag}$These authors contributed equally to this paper.}, Rafael~G.~Pires$^{\dag}$}
    \IEEEauthorblockA{
      Department of Computing\\
      S\~ao Paulo State University\\
      Bauru, Brazil\\
      \{danielfssantos1, rafapires\}@gmail.com}
  \and
	\IEEEauthorblockN{Danilo~Colombo}
	\IEEEauthorblockA{%
	  Cenpes\\
	  Petroleo Brasileiro S.A. - Petrobras\\
	  Rio de Janeiro - RJ, Brazil\\
	  colombo.danilo@petrobras.com.br
  \and
	\IEEEauthorblockN{Jo\~{a}o~P.~Papa}
	\IEEEauthorblockA{%
      Department of Computing\\
      S\~ao Paulo State University\\
      Bauru, Brazil\\
      joao.papa@unesp.br}}}

\else
  \author{Sibgrapi paper ID: 13 \\ }
\fi

\IEEEoverridecommandlockouts
\begin{document}

\maketitle

\begin{abstract}
Video segmentation consists of a frame-by-frame selection process of meaningful areas related to foreground moving objects. Some applications include traffic monitoring, human tracking, action recognition, efficient video surveillance, and anomaly detection. In these applications, it is not rare to face challenges such as abrupt changes in weather conditions, illumination issues, shadows, subtle dynamic background motions, and also camouflage effects. In this work, we address such shortcomings by proposing a novel deep learning video segmentation approach that incorporates residual information into the foreground detection learning process. The main goal is to provide a method capable of generating an accurate foreground detection given a grayscale video. Experiments conducted on the Change Detection 2014 and on the private dataset PetrobrasROUTES from Petrobras support the effectiveness of the proposed approach concerning some state-of-the-art video segmentation techniques, with overall F-measures of $\mathbf{0.9535}$ and $\mathbf{0.9636}$ in the Change Detection 2014 and PetrobrasROUTES datasets, respectively. Such a result places the proposed technique amongst the top 3 state-of-the-art video segmentation methods, besides comprising approximately seven times less parameters than its top one counterpart.
\end{abstract}


\IEEEpeerreviewmaketitle

\section{Introduction}
\label{s.introduction}

\IEEEPARstart{V}{ideo} segmentation refers to the process of highlighting some specific video image parts that belong to regions of interest, mostly associated to moving objects. Such a task is pretty much complicated to be solved in computer vision, presenting a great number of challenging situations that need to be considered such as extreme weather conditions, camera motion, subtle illumination changes, shadows cast by foreground objects, dynamic background motion, and camouflage. Therefore, addressing these challenges is crucial for the correct functioning of such a variety of computer vision applications including traffic monitoring~\cite{kato2002hmm}, human tracking~\cite{zhou2005real}, action recognition~\cite{ji20123d}, efficient video surveillance~\cite{brutzer2011evaluation}, and anomaly detection~\cite{chandola2009anomaly}. 

In the last decades, many non-learning and learning-dependent techniques have been developed to deal with the video segmentation problem. Amongst the non-learning dependent techniques we can highlight simple background subtraction~\cite{sakkos2018end,braham2016deep}, statistical~\cite{lanza2011statistical,varadarajan2013spatial,pulgarin2016gmm} and fuzzy models~\cite{bouwmans2012background}, subspace learning approaches~\cite{farcas2012background}, and robust Principal Component Analysis-based models~\cite{javed2015combining}. Amongst the learning-dependent techniques, we shall cite Quintana and Murguia~\cite{ramirez2015self} that proposed a bio-inspired neural system based on Self-organizing Maps and Cellular Neural Networks, called SOM-CNN, to detect dynamic objects in normal and complex scenarios. We can also refer to the work of Schofield et al.~\cite{schofield1996system}, that dealt with the problem of people segmentation and counting using a three-stage process: image pre-processing, background identification, and object search. Their method was designed to provide accurate counts, even when the background scene was allowed to vary.

Convolutional Neural Networks (CNNs) have gained quite an attention mostly because of their efficiency in solving tasks involving non-structured data~\cite{simonyan2014very}, as well as learning translational invariant properties, which is a key point for dealing with background motion detection. Learning-based video segmentation techniques that make use of CNNs are frequently emerging, such as the semi-automatic method for segmenting foreground moving objects proposed by Wang et al.~\cite{wang2017interactive}, which consists in two main objectives: (i) to produce segmentation maps sufficiently accurate to be used as a ground truth, and (ii) to avoid, as much as possible, user interventions. Another example concerns some state-of-the-art video segmentation techniques such as FgSegNet\_S and FgSegNet\_M~\cite{segnet2018foreground}, which are characterized mainly for being robust deep convolutional autoencoder networks that can be trained in an end-to-end model using a few video frames.

Some years ago, the concept of ``residual learning" arose to highlight the importance of considering skip connections to avoid a variety of deep network problems, such as vanishing gradients and overfitting. In this paper, we proposed a robust deep learning video segmentation technique that consists in a cascade CNN model that incorporates residual information~\cite{zhang2017beyond,he2016deep,pang2017cascade} into the learning process for foreground object detection. To the best of our knowledge, such an approach has never been investigated in the video segmentation domain. Experiments conducted in the public Change Detection 2014 (CD2014) and in the private PetrobrasROUTES datasets support the effectiveness of the proposed approach when detecting changes in indoor and outdoor camera-captured videos.

\begin{figure*}[!htb]
\centering
\includegraphics[width=4in,height=4in,keepaspectratio]{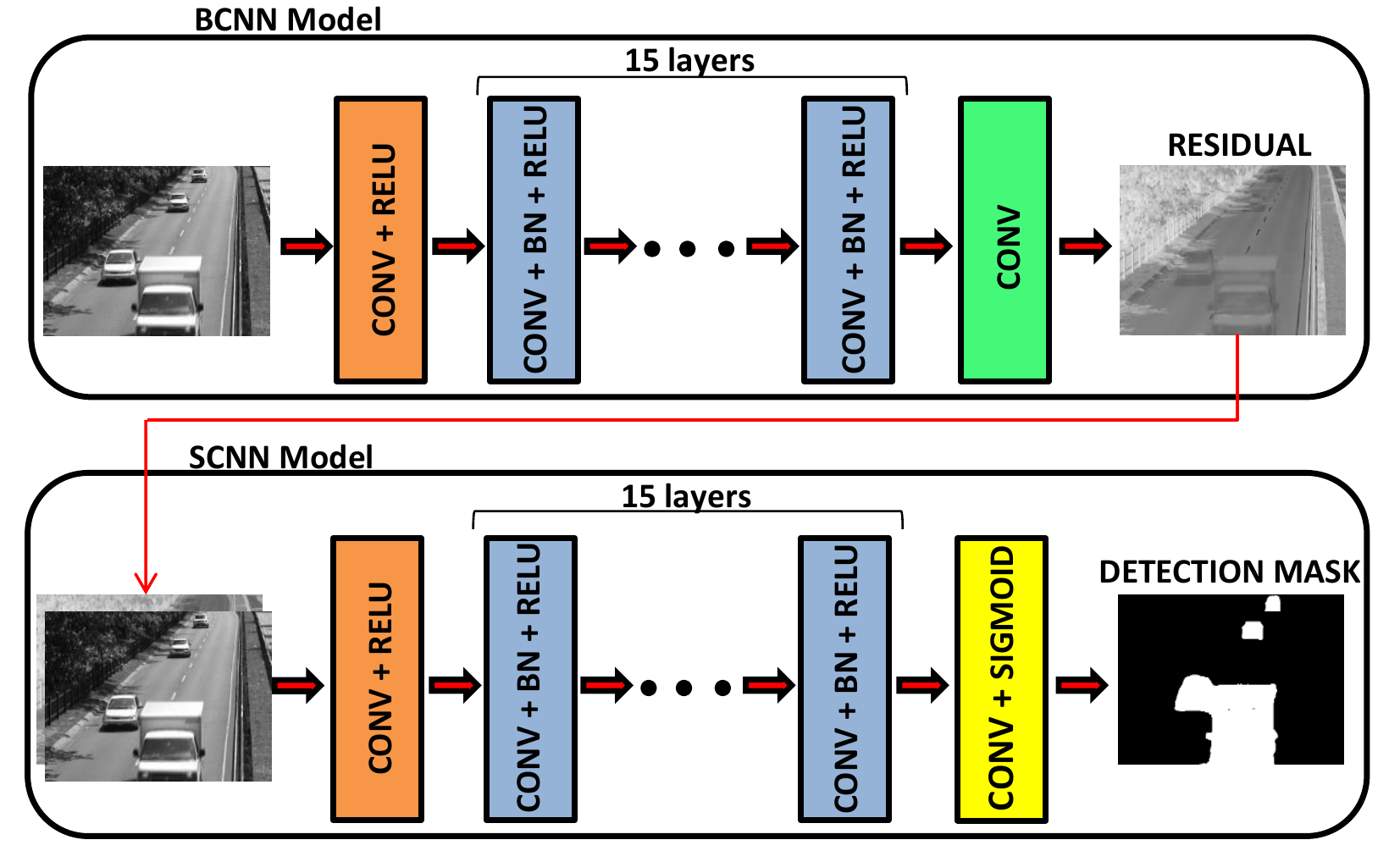}
  \caption{Architecture of the proposed CRCNN approach.}
  \label{f.crcnn_arch}
\end{figure*} 
\section{Proposed approach}
\label{s.prop_approach}


In this work, we proposed a novel approach named ``Cascade-Residual Convolutional Neural Network" (CRCNN) for video segmentation purposes, which was highly influenced by the works of Zhang et al.~\cite{zhang2017beyond} concerning non-blind denoising using residual learning, and Wang et al.~\cite{wang2017interactive} regarding segmentation issues. Besides, we also considered concepts from the work of Zhang et al.~\cite{zhang2018ffdnet} with respect to blind denoising learning. Figure~\ref{f.crcnn_arch} depicts the proposed approach.

The deep learning video segmentation pipeline adopted in this paper is the same one proposed by Lim and Keles~\cite{segnet2018foreground}, and consists in three main steps: (i) to annotate foreground objects in a small subset of frames collected from the video of interest, (ii) to train the CRCNN model in a supervised fashion, and (iii) to further apply the trained model over each image frame extracted from the video of interest to generate its correspondent binary foreground detection mask. 

The proposed CRCNN uses a two-stage video processing strategy. In the first step, given a grayscale version of the video image frame to be segmented, a first deep residual CNN, hereinafter called ``Background Convolutional Neural Network" (BCNN), is used to generate the correspondent residual map. Further,  this output combined with its residual map is presented to a second CNN, named ``Segmentation Convolution Neural Network" (SCNN), which is then used to generate the foreground detection mask. The next sections describe more details about the proposed CRCNN model.

\subsection{Background Convolution Neural Network}
\label{ss.proposed_approach_backcnn}

The proposed Background Convolutional Neural Network learns how to infer, given an input grayscale video frame, which parts do not correspond to the background areas. To create a robust background image, we used the approach proposed by Bevilacqua~\cite{bevilacqua2005simple} to allow BCNN modeling the input non-background information in the form of a residual map.

The BCNN training step consists of two phases: the first one takes an interval $I = \{\beta_{1}, \beta_{2}, ..., \beta_{n}\}$ of consecutive frames from the video and uses it to calculate the \emph{deterministic background image}, which stands for an image $b$ that represents the median of such an interval.  The second phase consists of minimizing the mean square error between the deterministic background image and the \emph{approximated background image} $a$, which is represented as follows:

\begin{equation}
\label{e.approximated_background}
a = \sigma(f - BCNN(f; \Theta_{1})),
\end{equation}
where $\sigma(\cdot)$ stands for the logistic-sigmoid function, $f$ denotes the input image normalized between $[0, 1]$, $\Theta_{1}$ refers to the BCNN trainable parameters, and $BCNN(\cdot;\cdot)$ refers to the residual map learned during the training process. In light of that, the BCNN training process aims at minimizing the following equation:

\begin{equation}
\label{e.backcnn_cost}
L_{B}(b, f; \Theta_{1}) = \frac{1}{2m}\sum_{i=1}^m||b_{i} - a_{i} ||_F^2,
\end{equation}
where $m$ stands for the number of training samples and $||\cdot||_F^2$ represents the Frobenius norm. Notice that we employed a patch-based methodology, where $b_i$ and $a_i$ denote the $i^{th}$ patch extracted from images $b$ and $a$, respectively.  

\subsection{Segmentation Convolutional Neural Network}
\label{ss.proposed_approach_segcnn}

The Segmentation Convolutional Neural Network learns how to detect foreground objects present in the video frames. For such purpose, it uses the information provided by the image frames and also their residual maps. The detections are presented in the form of binary images, in which white pixels correspond to the foreground object locations.
     
The SCNN training process differs from the BCNN one in basically two points: (i) the network input, that is composed of a concatenation between the grayscale image frame and its residual map counterpart (i.e., the output generated as a result of forward propagating the grayscale image through the trained BCNN), and (ii) the training process, which aims at minimizing the average binary cross-entropy measured between the network output and the ground-truth binary detection mask. Such an image corresponds to the pre-annotated true foreground objects present in the grayscale input image. Therefore, the SCNN training process aims at minimizing the following equation:    

\begin{equation}
\label{e.segcnn_cost}
L_{S}(g, c; \Theta_{2}) = -\frac{1}{m}\sum_{i=1}^m[g_{i}\log(\hat{g}_{i}) + (1 - g_{i})\log(1 - \hat{g}_{i})],
\end{equation}  
where
\begin{equation}
\label{e.segcnn_cost_p1}
\hat{g} = SCNN(c; \Theta_{2}).
\end{equation} 
Notice that $g$ is the ground-truth pre-annotated binary mask, $\Theta_{2}$ stands for the SCNN trainable parameters and $c$ indicates the SCNN depth concatenation input between $f$ and its residual map. Besides, $g_i$ and $\hat{g}_i$ denote the $i^{th}$ patch extracted from images $g$ and $\hat{g}$, respectively.

\subsection{Cascade Residual Convolutional Neural Network}
\label{ss.proposed_approach_paramscnn}

As depicted in Figure~\ref{f.crcnn_arch}, the proposed Cascade Residual Convolutional Neural Network is composed of two main models, i.e., the BCNN and SCNN, which are connected by the residual map generated by the former network. Table~\ref{t.arch_table} presents a summary of the CRCNN configuration parameters, where the dimensions of the convolution kernels are represented by three-dimensional vectors. The first and second dimensions represent the kernel width and height, respectively, and the third dimension denotes the number of outputs that will be generated after the convolution step. 
 
\begin{table}[htbp]
\renewcommand\arraystretch{1.4}
\centering
\caption{CRCNN architecture specification. The table uses the same color codes as in Figure~\ref{f.crcnn_arch} to represent the different CRCNN layers.}
\scalebox{1.05}{
\begin{tabular}{|c|c|c|c|c|}
\hline
\textbf{}                    & \textbf{Orange} & \textbf{Blue} & \textbf{Green} & \textbf{Yellow} \\ \hline
\textbf{Kernel szs.}         & 3 x 3 x 64      & 3 x 3 x 64    & 3 x 3 x 1      & 3 x 3 x 1       \\ \hline
\textbf{Activation}          & ReLU            & ReLU          & Linear         & Sigmoid         \\ \hline
\textbf{Batch norm.} & No              & Yes           & No             & No              \\ \hline
\end{tabular}}
\label{t.arch_table}
\end{table}

Three different kinds of activations were used, i.e., a linear function (applied to the convolution layers only), a rectified linear unit (ReLU), and a sigmoid function. Batch normalization~\cite{ioffe2015batch} was also applied in all $15$ layers, placed at the middle of the BCNN and SCNN models, i.e., before the application of the ReLU activation function. 
\section{Methodology}
\label{s.methodology}

In this section, we present the methodology used to train and evaluate the proposed CRCNN model. For the sake of clarification, we divided the section into three parts: (i) Section~\ref{ss.datasets} presents all the relevant information about the datasets used in this work, (ii) Section~\ref{ss.sub_train_proc} details the CRCNN training process, and (iii) Section~\ref{ss.sub_evaluation_proc} discusses the detection and evaluation procedures.       

\subsection{Datasets}
\label{ss.datasets}

\subsubsection{Change Detection Dataset 2014}
\label{sss.change_detection}

The Change Detection Dataset 2014 (CD2014) is a large and freely available dataset of videos collected from different realistic, camera-captured, and challenging scenarios~\cite{CD2014dataset}. Such a dataset contains $11$ video categories with $4$ to $6$ video sequences each, as presented in Table~\ref{t.cd2014_categories}.
   
\begin{table}[htbp]
\renewcommand\arraystretch{1.4}
\centering
\caption{CD2014 dataset specification.}
\scalebox{1.2}{
\begin{tabular}{|c|c|c|}
\hline
\textbf{Category}          & \textbf{Qnt. Videos} & \textbf{Qnt. Frames} \\ \hline
Baseline                   & 4                    & 6,049                 \\ \hline
Dynamic Background         & 6                    & 18,871                \\ \hline
Camera Jitter              & 4                    & 6,420                 \\ \hline
Intermitt. Obj. Motion & 6                    & 18,650                \\ \hline
Shadow                     & 6                    & 16,949                \\ \hline
Thermal                    & 5                    & 21,100                \\ \hline
Bad Weather                & 4                    & 20,900                \\ \hline
Low Framerate              & 4                    & 9,400                 \\ \hline
Night Videos               & 6                    & 16,609                \\ \hline
PTZ                        & 4                    & 8,630                 \\ \hline
Turbulence                 & 4                    & 15,700                \\ \hline
\textbf{Total}             & \textbf{53}          & \textbf{159,278}      \\ \hline
\end{tabular}}
\label{t.cd2014_categories}
\end{table} 

The CD2014 categories include:

\begin{itemize}
\item \textbf{Baseline}: combines mild challenges present in Dynamic Background, Camera Jitter, Intermittent Object Motion, and Shadow categories. It serves mainly as a starting point to adjust the segmentation technique.
 
\item \textbf{Dynamic Background}: includes scenes with so much background motion, e.g., cars and trucks passing in front of a tree shaken.

\item \textbf{Camera Jitter}: contains indoor and outdoor videos captured by unstable video devices, for example vibrating cameras.  
 
\item \textbf{Intermittent Object Motion}: contains objects that move and then stop for a short while producing ``ghosting" artefacts.

\item \textbf{Shadow}: indoor and outdoor videos containing objects surrounding by a strong shadow that could be miss detected as real moving objects.

\item \textbf{Thermal}: videos that have been captured by far-infrared cameras.

\item \textbf{Bad Weather}: includes outdoor videos captured from challenging winter weather conditions, e.g., snow storms, and fog.

\item \textbf{Low Framerate}: videos captured varying frame-rates between $0.17$fps and $1$fps.

\item \textbf{PTZ}: videos captured by pan-tilt-zoom cameras.

\item \textbf{Turbulence}: outdoor videos that show air turbulence caused by rising heat.   
\end{itemize}

\subsubsection{PetrobrasROUTES}
\label{sss.petrobras}

The PetrobrasROUTES is a private dataset which consists of $281$ high-resolution color images collected from an indoor Petrobras\footnote{Petrobras is a publicly-held company on an integrated basis and specialized in the oil, natural gas, and energy industry~\cite{Petrobrasdataset}.} workspace. The main challenge of such dataset regards the detection of objects obstructing escape routes.

\subsection{Training procedure}
\label{ss.sub_train_proc}

To train the proposed CRCNN model over CD2014 dataset, we employed the following protocol: 

\begin{enumerate}
\item to select $300$ color images and their $300$ correspondent binary images, which were ground-truth manually annotated.
\item to convert the $300$ color images to their grayscale versions and use the first $100$ images to calculate the deterministic background.
\item to normalize the remaining $200$ grayscale images\footnote{We used the same set of training images from~\cite{Segnetv2} to train the proposed CRCNN model.} by subtracting them the average grayscale value.
\item to subdivide the images into small patches using $50\%$ to $75\%$ of overlap depending on the image height and width dimensions\footnote{The sizes and the overlapping rates were empirically defined taking into account the usage of larger overlaps for small images and dimensions of the patches limited to $50$ pixels (i.e., for both height and width). Besides avoid much slowness during the training process, the usage of patches works like a natural data augmentation that prevents deep learning issues such as overfitting.}.
\item to subdivide the deterministic background image into small patches and then replicate them so that every input grayscale patch has its deterministic background grayscale patch counterpart.
\end{enumerate}

We employed the following protocol to train the proposed CRCNN model over PetrobrasROUTES dataset:

\begin{enumerate}
\item to select $51$ color images and their $51$ correspondent binary images, which were ground-truth manually annotated.
\item to convert the $51$ color images to their grayscale versions and use one of them as the deterministic background.
\item to normalize the remaining $50$ images by subtracting them the average grayscale value.
\item to subdivide the training and deterministic background images following same steps 4) and 5) from CD2014 dataset training protocol.
\end{enumerate} 

The BCNN and SCNN models were trained using the Adam method~\cite{kingma2014adam} by a maximum of $50$ epochs\footnote{Depending on the training process convergence the maximum epoch value can be less than 50.} using a learning rate\footnote{The initial value is reduced by a factor of $0.1$ every time the loss function hits a plateau.} of $0.001$ and batches of size $128$. We trained the BCNN and SCNN models with $80\%$ of the patches, and used the remaining $20\%$ to evaluate the convergence of the training process.

\subsection{Evaluation procedure}
\label{ss.sub_evaluation_proc} 

The evaluation process consists in to apply the trained CRCNN model over each video test image as follows:

\begin{itemize}
\item \textbf{Deep Segmentation}: such a step consists in first forward propagating the test images through the trained BCNN model, and further using the residual input (i.e., the input grayscale image and its residual counterpart) to feed the trained SCNN. Later, we binarized\footnote{In the majority of the experiments, the best threshold value was set to $0.8$, but in some rare cases, it has been set to $0.6$.} the SCNN probabilistic output.

\item \textbf{Misclassification Rate}: in such a step, we calculated the number of correct and incorrect detections encoded by the True Positives (TPs), i..e, the number of pixels correctly classified as foreground, the True Negatives (TNs), i.e., the number of pixels correctly classified as background, the False Positives (FPs), i.e., the number of background pixels incorrectly classified as foreground, and the False Negatives (FNs), i.e., the number of foreground pixels incorrectly classified as background.     

\item \textbf{Detection Measurements}: in such a step, the classification rates are combined into four different measures that provide a more clever way to measure the robustness of the proposed CRCNN model. Those measures are computed as follows: 

\begin{equation}\label{equ.precision}
Precision = \frac{TP}{TP+FP},
\end{equation}
\\*
\begin{equation}\label{equ.recall}
Recall = \frac{TP}{TP+FN},
\end{equation}
\\*
\begin{equation}\label{equ.fmeasure}
F-measure = 2.0 \times \frac{Recall \times Precision}{Recall + Precision},
\end{equation}
\\*
\noindent and
\\*
\begin{equation}\label{equ.pwc}
PWC = 100.0 \times \frac{FN + FP}{TP + FP + FN + TN} \\[15pt]
\end{equation}
\end{itemize}
where $PWC$ denotes the percentage of wrong classifications.

\section{Experimental Section}
\label{s.experimental_section}

\begin{table*}[hbt!]
\centering
\renewcommand\arraystretch{1.5}
\setlength{\tabcolsep}{.58em}
\caption{A Comparison of F-measure results of 11 categories from CD2014 dataset}
\scalebox{1.1}{
\begin{tabular}{crrrrrrrrrrrr}
\toprule 
Methods & Baseline & C.Jitter & B.Waet & Dyn.Bg. & Int.Obj. & L.Frame & N.Videos & PanTZ   & Shadow & Thermal & Turbul. &  Overall  \\ 
\midrule
FgSegNet\_S & \textbf{0.9980}   & \textbf{0.9951}      & \textbf{0.9902}    & \textbf{0.9902}   & \textbf{0.9942}       & \textbf{0.9511}      & \textbf{0.9837}     & \textbf{0.9837} & \textbf{0.9967} & \textbf{0.9945}  & \textbf{0.9796}  & \textbf{0.9878} \\
FgSegNet\_M & \textbf{0.9975}   & \textbf{0.9945}      & \textbf{0.9838}    & \textbf{0.9838}   & \textbf{0.9933}       & \textbf{0.9558}      & \textbf{0.9779}     & \textbf{0.9779} & \textbf{0.9954} & \textbf{0.9923}  & \textbf{0.9776}  & \textbf{0.9865} \\
CRCNN       & \textbf{0.9919}   & \textbf{0.9799}      & \textbf{0.9569}    & \textbf{0.9687}   & \textbf{0.9755}       & 0.8498      & \textbf{0.9388}     & \textbf{0.8967} & \textbf{0.9852} & \textbf{0.9818}  & \textbf{0.9637}  & \textbf{0.9535} \\
Cascade     & 0.9786   & 0.9758      & 0.9451    & 0.9451   & 0.8505       & \textbf{0.8804}      & 0.8926     & 0.8926 & 0.9593 & 0.8958  & 0.9215  & 0.9272 \\
DeepBS      & 0.9580   & 0.8990      & 0.8647    & 0.8647   & 0.6097       & 0.5900      & 0.6359     & 0.6359 & 0.9304 & 0.7583  & 0.8993  & 0.7593 \\
IUTIS-5     & 0.9567   & 0.8332      & 0.8289    & 0.8289   & 0.7296       & 0.7911      & 0.5132     & 0.5132 & 0.9084 & 0.8303  & 0.8507  & 0.7820 \\
PAWCS       & 0.9397   & 0.8137      & 0.8059    & 0.8059   & 0.7764       & 0.6433      & 0.4171     & 0.4171 & 0.8934 & 0.8324  & 0.7667  & 0.7477 \\
SuBSENSE    & 0.9503   & 0.8152      & 0.8594    & 0.8594   & 0.6569       & 0.6594      & 0.4918     & 0.4918 & 0.8986 & 0.8171  & 0.8423  & 0.7453 \\ \bottomrule
\end{tabular}}
\label{t.fmeasures}
\end{table*}  

In this section, we present the experimental results regarding the methodology described earlier considering each dataset. 

\subsection{Results over CD2014 Dataset}
\label{ss.results_cd2014}

Table~\ref{t.fmeasures} presents the overall and per-category F-measure values. One can 
observe the proposed CRCNN model overcomes the supervised learning methods Cascade~\cite{wang2017interactive} and DeepBS~\cite{babaee2018deep}, being also more accurate in comparison to the non-learning-based techniques, i.e., SuBSENSE~\cite{subsense}, IUTIS-5~\cite{bianco2017combination}, and PAWCS~\cite{st2015self}. According to Table~\ref{t.fmeasures}, the proposed technique achieved results that are pretty much close to the state-of-the-art ones, as one can notice in the categories ``Baseline", ``Camera Jitter", and ``Shadow", where CRCNN results are quite similar to the FgSegNet\_S and FgSegNet\_M~\cite{segnet2018foreground} techniques, with F-measure differences of only $0.01$ (approximately). Notice the proposed approach comprises $1,112,770$ parameters, which turns out to be a more compact architecture with respect to FgSegNet\_S, which has $7,622,465$ parameters.  

Table~\ref{t.overall_results} highlights the robustness of the proposed CRCNN model, placing it as the third best approach concerning the measures used in this work, only behind FgSegNet\_S and FgSegNet\_M models. Also, the precision results differ from FgSegNet\_S and FgSegNet\_M by around $0.02$, while recall values differ by around $0.03$.

\begin{table}[hbt!]
\renewcommand{\arraystretch}{1.5}
\centering
\caption{Comparison of precision, recall and PWC overall results from CD2014 dataset.}
\scalebox{1.15}{
\begin{tabular}{cccc}
\hline
Methods     & Avg. Precision  & Avg. Recall     & Avg. PWC        \\ \hline
FgSegNet\_S & \textbf{0.9751} & \textbf{0.9896} & \textbf{0.0461} \\
FgSegNet\_M & \textbf{0.9758} & \textbf{0.9836} & \textbf{0.0559} \\
CRCNN       & \textbf{0.9604} & \textbf{0.9602} & \textbf{0.1348} \\
Cascade     & 0.8997          & 0.9506          & 0.4052          \\
DeepBS      & 0.8332          & 0.7545          & 1.9920          \\
IUTIS-5     & 0.8087          & 0.7849          & 1.1986          \\
PAWCS       & 0.7857          & 0.7718          & 1.1992          \\
SuBSENSE    & 0.7509          & 0.8124          & 1.6780          \\ \hline
\end{tabular}}
\label{t.overall_results}
\end{table}

\begin{figure}[htb!]
\setlength{\tabcolsep}{2pt} 
\renewcommand{\arraystretch}{1.1} 
\hspace*{.05cm}
\centerline{
\begin{tabular}{ccc}
{\small Night Videos}    &   {\small Shadow}  &   {\small Thermal}\\
   \includegraphics[width=2.8cm, height=2.8cm]{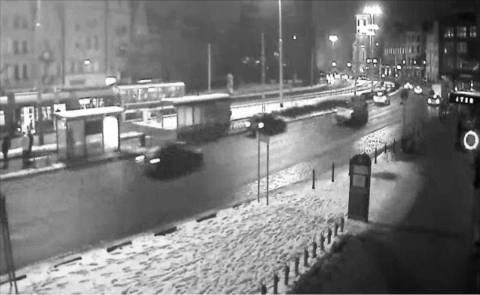}&
      \includegraphics[width=2.8cm, height=2.8cm]{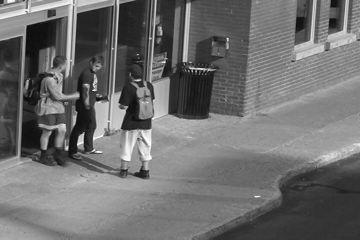}&
      \includegraphics[width=2.8cm, height=2.8cm]{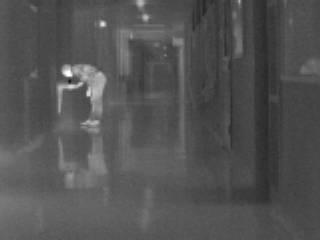}\\
      & (a) & \\
\end{tabular}}
\hspace*{.05cm}
\centerline{
\begin{tabular}{rccc}
      \includegraphics[width=2.8cm, height=2.8cm]{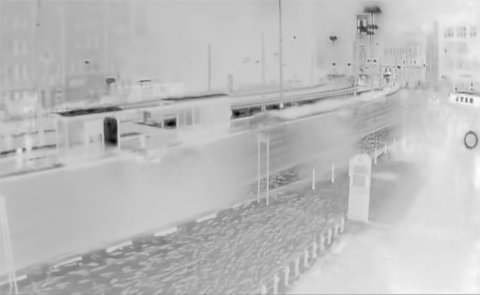}&
      \includegraphics[width=2.8cm, height=2.8cm]{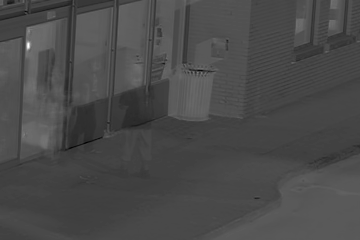}&
      \includegraphics[width=2.8cm, height=2.8cm]{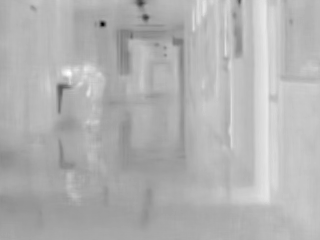}\\
      & (b) & \\
\end{tabular}}
\hspace*{.05cm}
\centerline{
\begin{tabular}{rccc}
      \includegraphics[width=2.8cm, height=2.8cm]{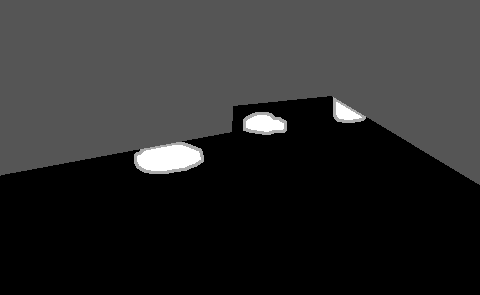}&
      \includegraphics[width=2.8cm, height=2.8cm]{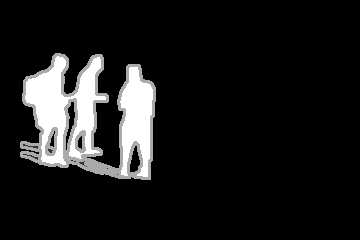}&
      \includegraphics[width=2.8cm, height=2.8cm]{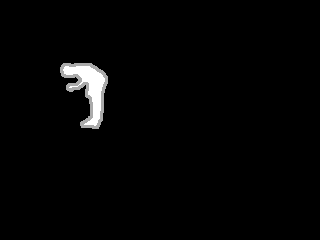}\\
      & (c) & \\
\end{tabular}}
\hspace*{.05cm}
\centerline{
\begin{tabular}{rccc}
      \includegraphics[width=2.8cm, height=2.8cm]{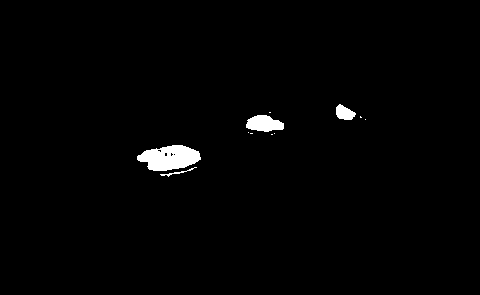}&
      \includegraphics[width=2.8cm, height=2.8cm]{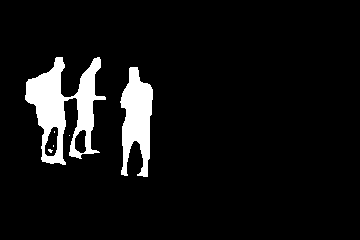}&
      \includegraphics[width=2.8cm, height=2.8cm]{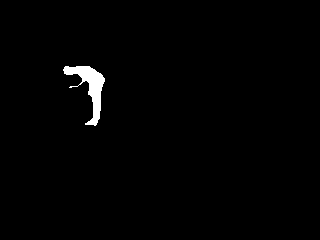}\\
     & (d) & \\
\end{tabular}}
\hspace*{.05cm}
\centerline{
\begin{tabular}{rccc}
      \includegraphics[width=2.8cm, height=2.8cm]{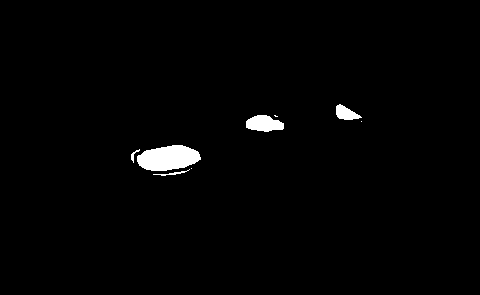}&
      \includegraphics[width=2.8cm, height=2.8cm]{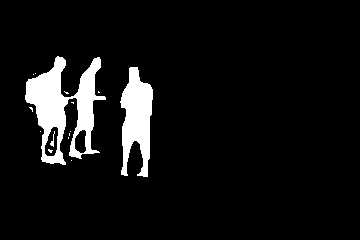}&
      \includegraphics[width=2.8cm, height=2.8cm]{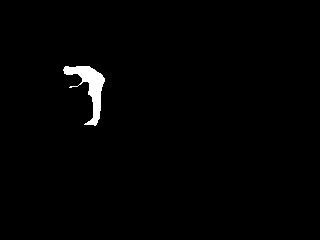}\\
     & (e) & \\
\end{tabular}}
\hspace*{.05cm}
\centerline{
\begin{tabular}{rccc}
      \includegraphics[width=2.8cm, height=2.8cm]{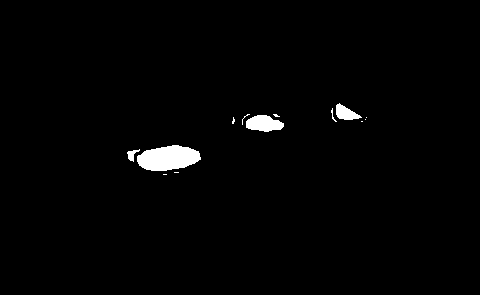}&
      \includegraphics[width=2.8cm, height=2.8cm]{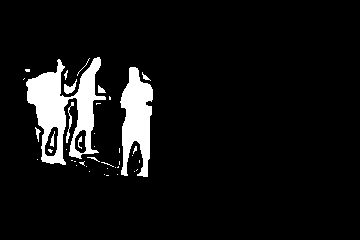}&
      \includegraphics[width=2.8cm, height=2.8cm]{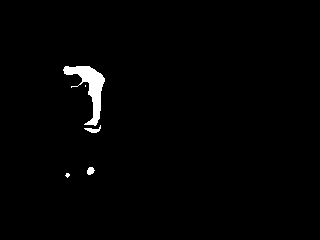}\\
      & (f) & \\
\end{tabular}}
\hspace*{.05cm}
\centerline{
\begin{tabular}{rccc}
      \includegraphics[width=2.8cm, height=2.8cm]{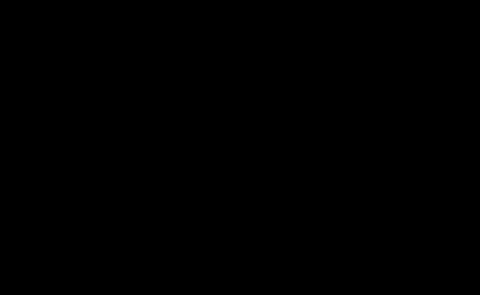}&
      \includegraphics[width=2.8cm, height=2.8cm]{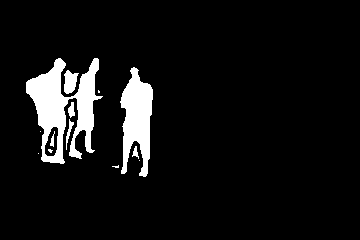}&
      \includegraphics[width=2.8cm, height=2.8cm]{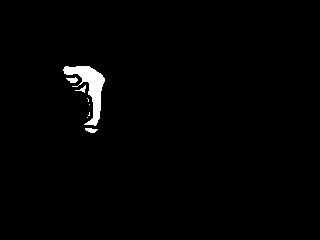}\\
      & (g) & \\
\end{tabular}}
\centering
\caption{Qualitative results considering the categories ``Night Videos", ``Shadow", and ``Thermal" from CD2014 dataset: (a) input grayscale frame, (b) residual maps, (c) ground-truth detection masks, results concerning (d) proposed CRCNN, (e) FgSegNet\_S, (f) Cascade, and (g) DeepBS models.}
\label{f.clean_noise_restoration}
\end{figure}

Additionally to the results presented in Tables~\ref{t.fmeasures} and~\ref{t.overall_results}, Figure~\ref{f.clean_noise_restoration} depicts three foreground detection masks, each one from a different category. Notice the gray-tone areas presented in Figure~\ref{f.clean_noise_restoration}c stand for regions that do not belong to the region of interest. From the ground-truth presented in Figure~\ref{f.clean_noise_restoration}c, one can observe the proposed CRCNN model produced a more accurate and precise detection binary masks for the ``Shadow" category (Figure~\ref{f.clean_noise_restoration}d). These results are better than the ones obtained by FgSegNet\_S, Cascade, and DeepBS techniques. Concerning the ``Thermal category", CRCNN outperformed Cascade and DeepBS models.      

One can also observe that CRCNN exhibited false negative detections to a greater extent when compared to FgSegNet\_S and Cascade techniques in the ``Night Videos" category. However, such a category poses a challenge to all compared methods either, since Cascade and FgSegNet\_S results exhibit false positive detections both, and DeepBS was not capable of detecting the moving cars in the video frame. A closer look at the second row from Figure~\ref{f.clean_noise_restoration} evidenced that BCNN model smoothed areas corresponding to foreground regions during its learning process. We hypothesized that such regions are used by SCNN during its learning process as a clue indicating which locations are the most probable to encode scene changes.


\subsection{Results over PetrobrasROUTES Dataset}
\label{ss.results_petrobras}

Table~\ref{t.overall_petro_results} presents the overall results comparing FgSegNet\_S with the proposed technique. One can 
observe the proposed CRCNN achieved the best results in almost all measures, with differences of around $0.1$ and $0.2$ in terms of recall and PWC, respectively.  

\begin{table}[hbt!]
\renewcommand{\arraystretch}{1.5}
\centering
\caption{Comparison of precision, recall and PWC overall results from PetrobrasROUTES dataset.}
\scalebox{1.15}{
\begin{tabular}{ccc}
\hline
Avg. Measures & FgSegNet\_S     & CRCNN           \\ \hline
F-measure     & 0.9221          & \textbf{0.9619}          \\
Precision     & \textbf{0.9770} & 0.9611          \\
Recall        & 0.8732          & \textbf{0.9627} \\
PWC           & 0.4287          & \textbf{0.2218} \\ \hline
\end{tabular}}
\label{t.overall_petro_results}
\end{table}

\begin{figure}[htb!]
\setlength{\tabcolsep}{2pt} 
\renewcommand{\arraystretch}{1.5} 
\vspace*{.45cm}
\centerline{
\begin{tabular}{cr}
   \includegraphics[width=8cm, height=3.5cm]{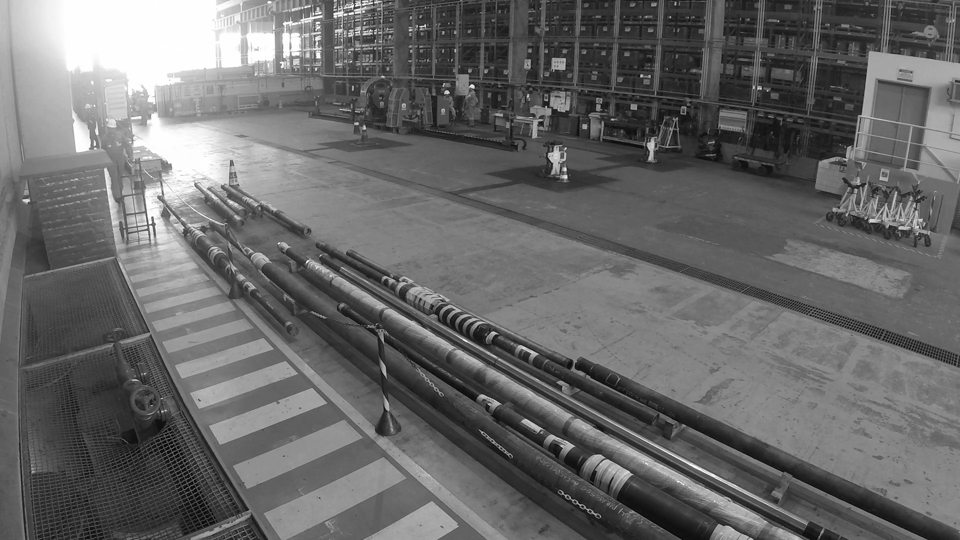}\\
   (a)\\
\end{tabular}}
\centerline{
\begin{tabular}{cr}
   \includegraphics[width=8cm, height=3.5cm]{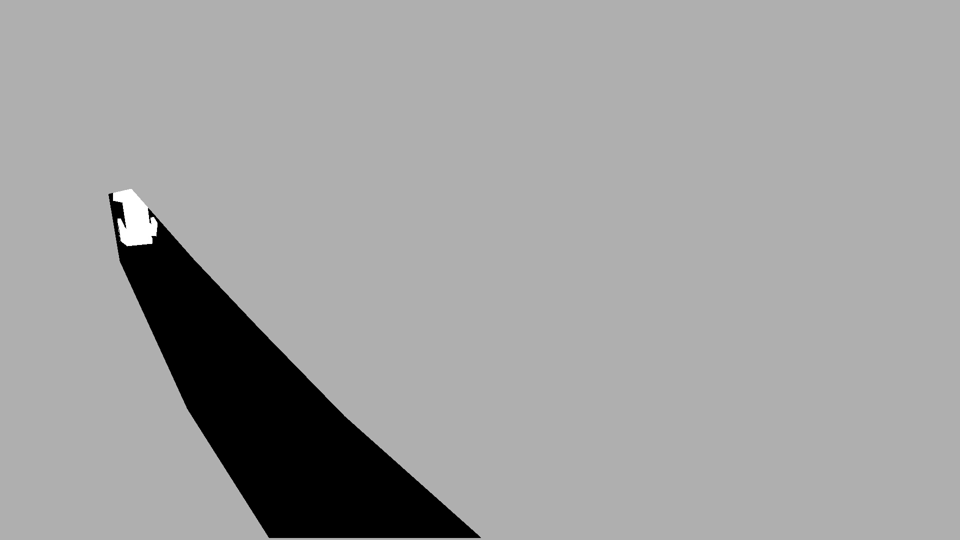}\\
   (b)\\
\end{tabular}}
\centerline{
\begin{tabular}{cr}
      \includegraphics[width=8cm, height=3.5cm]{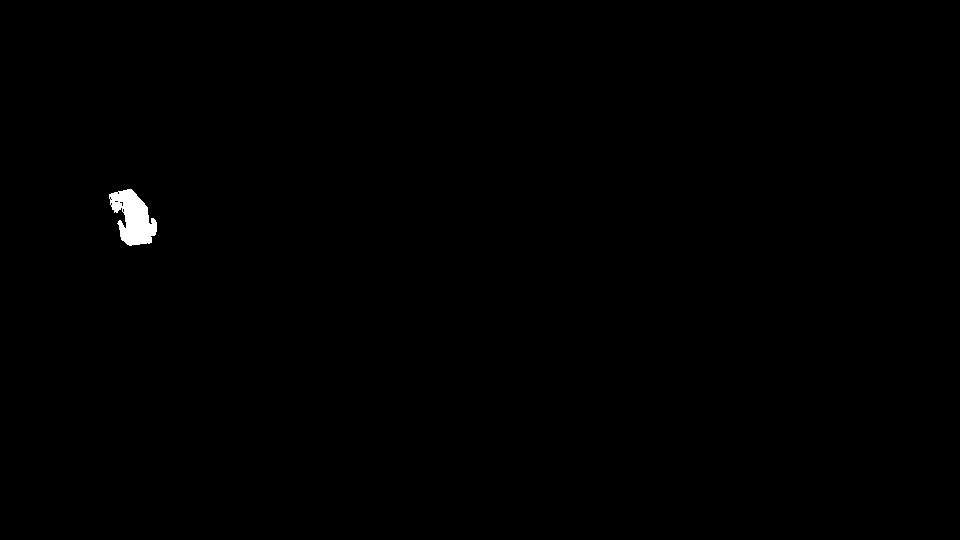}\\
      (c)\\
\end{tabular}}
\centerline{
\begin{tabular}{cr}
      \includegraphics[width=8cm, height=3.5cm]{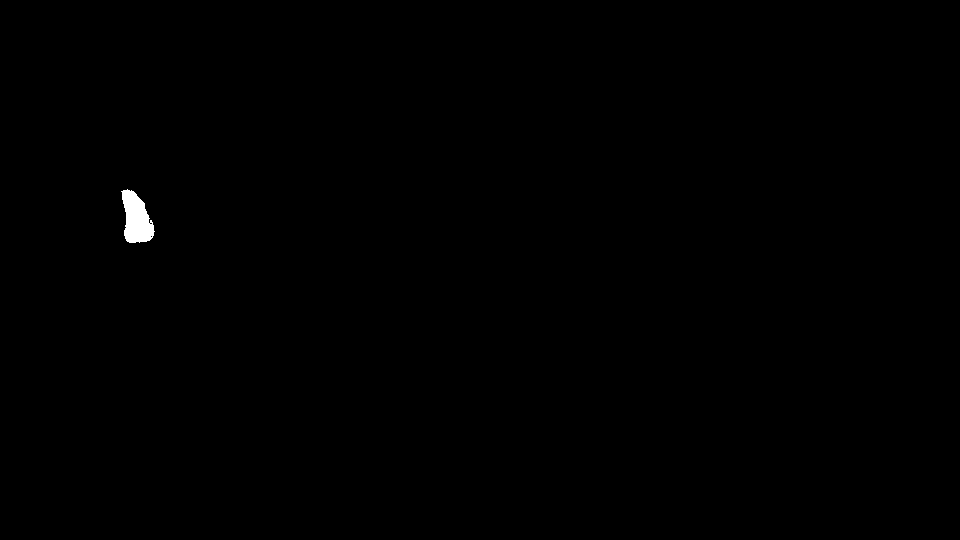}\\
      (d)\\
\end{tabular}}
\centering
\caption{Qualitative results considering an obstructed route video scene from PetrobrasROUTES dataset: (a) input grayscale frame, (b) ground-truth detection mask, and results concerning (c) CRCNN and (d) FgSegNet\_S techniques.}
\label{f.petro_segmentation}
\end{figure}

Additionally to the results presented in Table~\ref{t.overall_petro_results}, Figure~\ref{f.petro_segmentation}a depicts a video frame of a scape route containing an undesirable object. Regards its ground-truth (Figure~\ref{f.petro_segmentation}b), one can observe the proposed CRCNN has been more accurate than FgSegNet\_S (Figures~\ref{f.petro_segmentation}c and~\ref{f.petro_segmentation}d), with the detection results limited only to the object central region.  

\section{Conclusions}
\label{s.conclusions}

In this work, we proposed a novel cascade convolutional neural network which uses a residual learning strategy in an attempt to solve video segmentation problems. Regarding CD2014 dataset, the proposed CRCNN model achieved results close to the state-of-the-art ones, which were obtained by FgSegNet\_S and FgSegNet\_M techniques. Besides, the method was capable of overcoming two supervised learning methods and three other non-supervised segmentation techniques in terms of F-measure, precision, recall, and PWC overall results. Concerning PetrobrasROUTES dataset, the proposed CRCNN model outperformed the state-of-the-art FgSegNet\_S method in terms of F-measure, recall, and PWC overall results. Besides, we state that better results can be possibly achieved by fine-tuning the patch sizes. 

Regarding future works, we pretend to investigate the CRCNN behavior under such a fine-tuning process carefully, and also search for other possible ways to apply the residual learning in video segmentation tasks. As a starting point, we intend to investigate the usage of color images and possibly other CNN architecture configurations. 

\iffinal
\section*{Acknowledgment}
The authors are grateful to CNPq grants 307066/2017-7 and 427968/2018-6, FAPESP grants 2013/07375-0, 2014/12236-1 and 2016/19403-6, as well as Petrobras grant 2017/00285-6.
\fi

\bibliographystyle{IEEEtran}
\bibliography{ms}

\end{document}